\documentclass[letterpaper]{article} 
\usepackage{aaai2027}  
\usepackage[hyphens]{url}  
\usepackage{graphicx} 
\usepackage{natbib}  
\usepackage{caption} 
\usepackage{algorithm}
\usepackage{algorithmic}

\usepackage{newfloat}
\usepackage{listings}
\DeclareCaptionStyle{ruled}{labelfont=normalfont,labelsep=colon,strut=off} 
\floatstyle{ruled}
\newfloat{listing}{tb}{lst}{}
\floatname{listing}{Listing}

\usepackage{booktabs}
\usepackage{amsmath}
\usepackage{amssymb}

\nocopyright

\title{SepsisLens: Structure-Preserving Sequence Modelling for Decomposable Early Sepsis Warning}
\author{
    Yikun Ou,
    Wei Li}
\affiliations{
    School of Computer Science, University of Sydney\\
    Sydney, Australia
}

\begin{document}

\maketitle

\begin{abstract}
Early sepsis warning from ICU records can be cast as a structure-preserving prediction problem. A model needs to detect deterioration from irregular measurements while keeping each alert connected to the physiological signals that support it. Many temporal models fuse clinical variables into a patient-level representation, supporting scalar risk prediction but weakening the structure needed for clinical decomposition. We present SepsisLens, which preserves variable-indexed temporal states until risk composition. Observation-aware representations encode each variable's dynamics and measurement history, while a shared temporal encoder models each trajectory without collapsing the variable axis. The StructuredRiskHead composes multi-horizon risk from explicit variable-level and organ-level components. We evaluate SepsisLens on three public ICU cohorts and one private-hospital cohort under a common pre-onset protocol. SepsisLens achieves strong discrimination on all four cohorts and
lower alert burden at matched event recall on MIMIC-IV. Structural ablations support the design, while input-side masking shows that the ranked components reflect variables with greater influence on prediction.
\end{abstract}

\section{Introduction}
\label{sec:Introduction}

In the ICU, patients with emerging sepsis can deteriorate before the diagnosis is clear. A useful warning needs to detect this change in time to support intervention while remaining selective enough to limit false alarms. It also needs to show which physiological signals support the warning. A single risk score can raise concern, but the score is difficult to act on when detached from the signals behind it.

ICU electronic records make this warning task difficult. Clinical variables are measured at different frequencies in ICU care. Vital signs may be recorded often, laboratory tests may appear only intermittently, and some measurements may never be ordered. These observation patterns are clinically meaningful. The decision to measure a variable can reflect concern about the patient's condition. Sepsis adds another layer of structure. Organ dysfunction is central to its assessment, and SOFA summarises this dysfunction while retaining an organ-level account \cite{sepsis3,sofa}. The score is an aggregate value, but it remains linked to the organ systems that contribute to the overall assessment. A model that returns only a scalar risk score can lose this link between prediction and physiological organisation.

Learning-based sepsis predictors have improved several parts of early warning. Recurrent and temporal models learn patient-level risk from ICU trajectories \cite{futoma2017mgprnn,rosnati2021mgpatttcn}. Irregular-time models account for uneven measurements and missingness \cite{grud,shukla2021mtan,raindrop}. Attention-based models relate predictions to clinical variables \cite{choi2016retain}, while calculator-informed graph models integrate clinical scores into temporal prediction \cite{sepsiscalc}. These advances leave a structure-preservation problem unresolved. Many temporal predictors fuse clinical variables into a patient-level representation before variable-specific temporal states are formed. We call this design early fusion. Early fusion can support scalar discrimination, but it weakens the variable structure needed for decomposable warning. Once variables have been mixed into a global hidden state, the prediction path no longer retains an explicit variable axis. The model may still rank risk well. The resulting risk, however, is not naturally composed from variable-level and organ-level components. Post-hoc attribution can identify inputs associated with a prediction \cite{jain2019attention,rudin2019stop}. It does not make the prediction itself a structured composition of clinical components. Much prior work asks how a computed risk score can be explained. We ask how to preserve the clinical structure needed for risk to remain decomposable during prediction.

\begin{figure*}[t]
    \centering
    \includegraphics[width=0.93\textwidth]{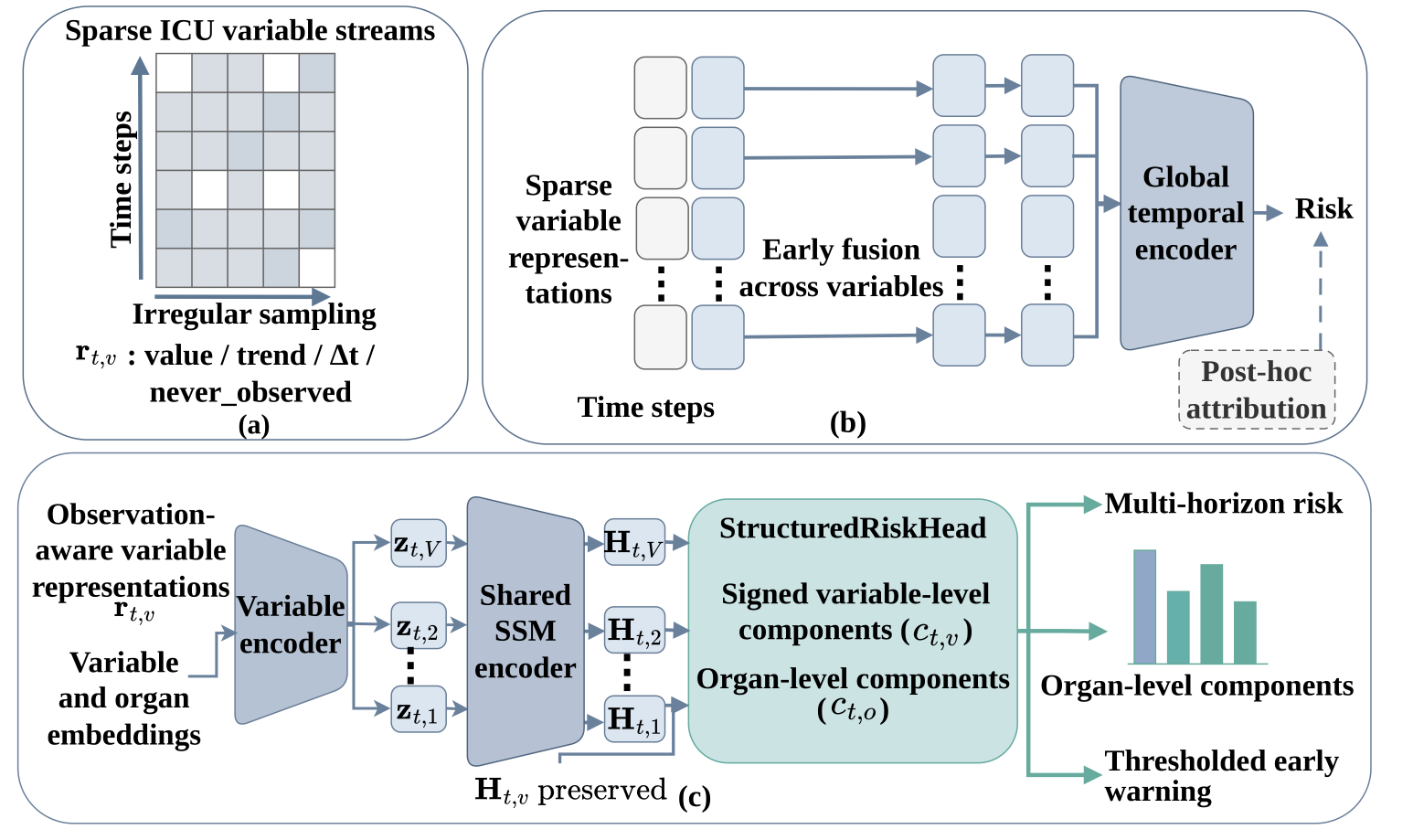}
    \caption{
        Motivation for SepsisLens.
        (a) Irregular ICU observations contain information about variable values, changes, and measurement history.
        (b) Early fusion collapses variable trajectories into a global representation without explicit variable- or organ-level components. Post-hoc attribution is applied outside the risk computation.
        (c) SepsisLens preserves variable-indexed temporal states until risk composition, producing multi-horizon risk with explicit variable- and organ-level components.
    }
    \label{fig:motivation}
\end{figure*}

SepsisLens addresses this gap by preserving variable-indexed temporal states until risk composition. Observation-aware variable representations retain value dynamics, measurement timing, and never-observed status. A shared temporal encoder models each variable trajectory while preserving the variable axis. The StructuredRiskHead computes signed variable-level components, aggregates them into organ-level components, and adds a global residual correction to each horizon-specific risk logit. Organ-dysfunction supervision aligns the organ-level components during training. This design places an explicit clinical component path inside the forward computation and retains a residual path for global correction. Figure~\ref{fig:motivation} summarises this structure-preserving path and contrasts it with early-fusion prediction and post-hoc attribution. We evaluate SepsisLens on three public ICU cohorts and one private-hospital cohort using a common pre-onset protocol. The evaluation examines predictive discrimination, alert burden at matched recall, the contribution of the structural design, and the functional relationship between the ranked components and prediction.

The main contributions are as follows:
\begin{itemize}
    \item We formulate ICU sepsis warning as a structure-preserving early-event prediction problem, where variable identity remains available until risk composition and is not removed by early fusion.
    
    \item We develop observation-aware variable representations and a variable-preserving temporal encoder that models each clinical variable as an identifiable temporal state until prediction.
    
    \item We introduce the StructuredRiskHead, which forms multi-horizon risk from signed variable-level components, organ-level aggregation, and a global residual correction, with organ-dysfunction supervision for component alignment.
    
    \item We evaluate SepsisLens on four ICU cohorts under a common pre-onset and validation-selected early-warning protocol. Results show strong discrimination on all cohorts, lower false alarm rates at matched event recall on MIMIC-IV, and functional relevance of structured components via ablations and input-side masking.
\end{itemize}

\section{Related Work}
\label{sec:Related_Work}

\paragraph{Early sepsis warning and irregular clinical time series.}
Early sepsis warning is often modelled as a clinical time-series prediction task. RETAIN \cite{choi2016retain} links predictions to visits and variables using reverse-time attention. MGP-RNN \cite{futoma2017mgprnn} uses Gaussian process interpolation before recurrent modelling, and MGP-AttTCN \cite{rosnati2021mgpatttcn} applies it before convolutional temporal modelling. GRU-D uses missingness and elapsed time in recurrent updates \cite{grud}. Multi-time attention networks model irregular observations with continuous-time attention \cite{shukla2021mtan}, and Raindrop uses graph-guided temporal structure \cite{raindrop}. Recent methods \cite{xiao2025xtsformer,liu2025timecheat,liu2026apn,zhang2024tpatchgnn} extend these ideas to asynchronous clinical events and more flexible representations of irregular time series. These methods improve prediction under sparse and uneven observations. They do not focus on preserving variable-indexed temporal states until risk composition. SepsisLens addresses this gap by keeping these states explicit until risk composition.

\paragraph{Structured clinical prediction.}
Clinical sepsis assessment is organised around organ dysfunction, as reflected in Sepsis-3 and SOFA \cite{sepsis3,sofa}. This structure motivates models that retain clinically meaningful components within prediction. Attention-based models can associate predictions with visits or variables \cite{choi2016retain}. Post-hoc attribution methods can support inspection \cite{jain2019attention,rudin2019stop}. Neither makes the risk score itself a structured composition of clinical components. SepsisCalc incorporates clinical calculators into a dynamic temporal graph for early sepsis prediction, allowing clinical scoring information to enter the prediction process \cite{sepsiscalc}. SepsisLens addresses a different representation problem. It preserves variable-indexed temporal states during sequence encoding. The forward pass forms explicit variable-level and organ-level components before adding a global residual correction.

\paragraph{Operational evaluation for early warning.}
External validation of a widely used sepsis prediction model reported substantial alert burden at a threshold that still missed many sepsis cases \cite{wong2021external}. Early-warning models need to be evaluated beyond ranking metrics. Models with similar AUROC can produce different alert burdens after thresholding. A model can achieve high event recall by alerting often, while a low false alarm rate may simply reflect missed deterioration. This motivates our use of validation-selected thresholds and threshold-sweep analyses to compare operating profiles without reducing the comparison to a single operating point. We report threshold-independent discrimination alongside event recall, lead time, and false alarm rate.

\section{Problem Formulation}
\label{sec:Problem_Formulation}
For each patient, we construct an hourly aligned clinical trajectory
with up to $T = 96$ bins, including up to 24 pre-ICU alignment bins,
and $V = 31$ clinical variables. Let $t=1,\ldots,T$ index timesteps and $a_t$ denote their aligned
clinical hours relative to ICU admission, with $a_t<0$ for pre-ICU bins. The input contains a value matrix $\mathbf{X}\in\mathbb{R}^{T\times V}$ and a cumulative observation-history mask $\mathbf{M}\in\{0,1\}^{T\times V}$, where $M_{t,v}=1$ indicates that variable $v$ has been observed at or before timestep $t$. Let $\mathcal{H}=\{3,6,12\}$ denote the prediction horizons in hours. Let $\tau$ denote the sepsis onset hour on the aligned time grid, with $\tau=\infty$ for patients who do not develop sepsis. At each timestep $t$ before onset, the target for horizon $h\in\mathcal{H}$ is
\begin{equation}
y_t^{(h)}
=
\mathbb{I}(a_t < \tau \leq a_t+h).
\end{equation}

where $\mathbb{I}(\cdot)$ is the indicator function. For non-sepsis patients, $\tau=\infty$ makes all horizon targets negative.

The model outputs risk estimates for all horizons using only the observed history up to timestep $t$:
\begin{equation}
\hat{\mathbf{y}}_t
=
f(\mathbf{X}_{1:t},\mathbf{M}_{1:t}),
\qquad
\hat{\mathbf{y}}_t
=
\left\{\hat{y}^{(h)}_t\right\}_{h\in\mathcal{H}},
\end{equation}
where $f$ is the learned prediction function and $\mathbf{X}_{1:t},\mathbf{M}_{1:t}$ denote the patient history available at prediction time.

The 6-hour horizon is the primary reporting horizon unless otherwise specified. Onset and post-onset timesteps are excluded from the early-warning loss. During training, time-aligned SOFA-derived organ dysfunction
labels provide auxiliary supervision for the organ-level
components. These labels are used during training
only. At inference, the model receives only the patient trajectory. The task is to learn a history-only prediction function that produces multi-horizon sepsis risk while retaining an explicit variable-level and organ-level component path within the forward computation. The resulting risk estimates are thresholded to produce warnings, which are evaluated using event recall, lead time, and false alarm rate.

\section{SepsisLens}

\begin{figure*}[t]
    \centering
    \includegraphics[width=0.97\textwidth]{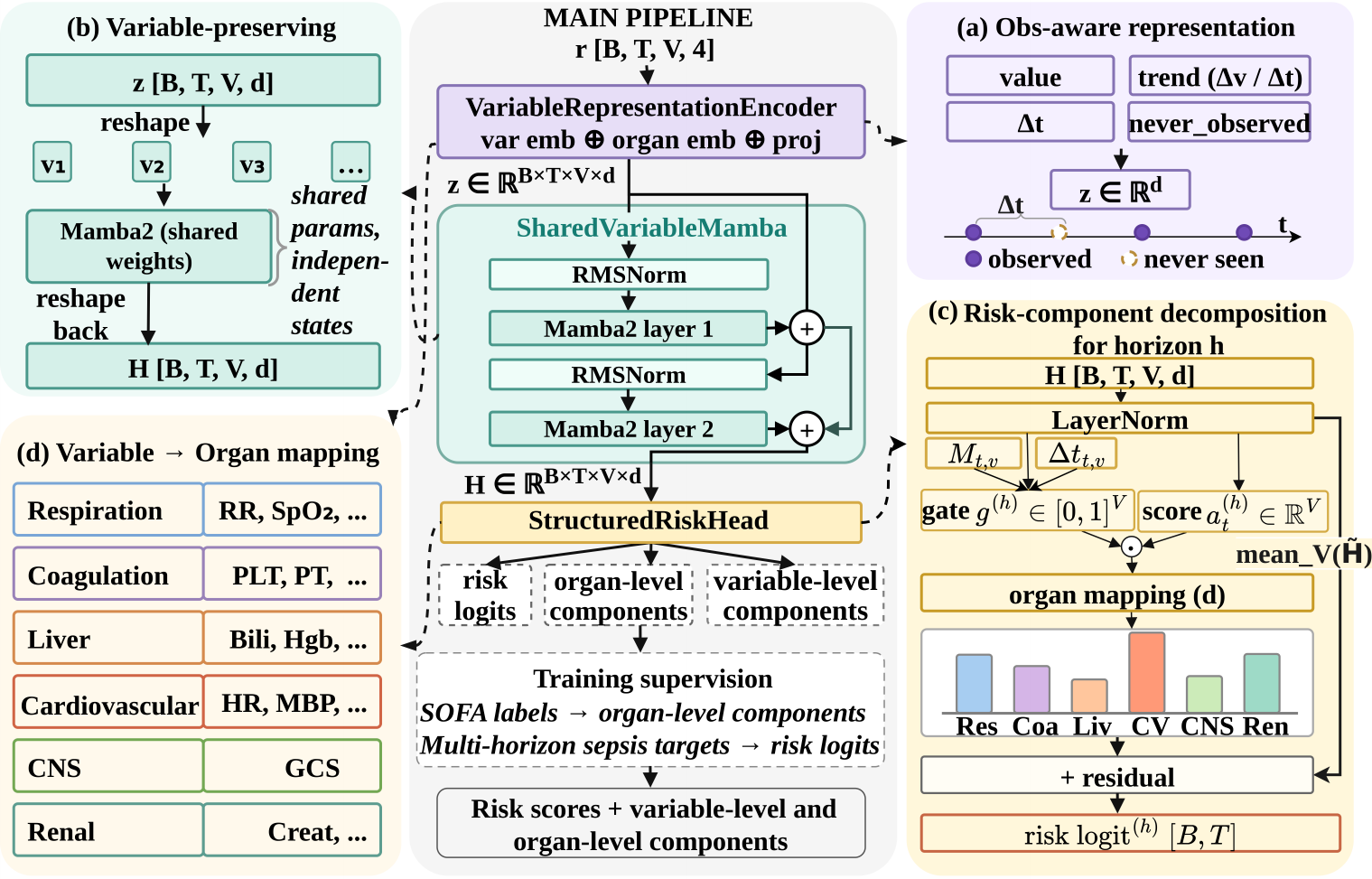}
    \caption{%
    SepsisLens architecture.
    (a) Observation-aware representations encode value, trend,
    elapsed time, and never-observed status for each clinical variable.
    (b) A shared temporal encoder models variable trajectories
    and preserves variable-indexed states
    $\mathbf{H}\in\mathbb{R}^{B\times T\times V\times d}$.
    (c) StructuredRiskHead computes gated variable-level components,
    aggregates them into organ-level components, and adds a global residual correction.
    (d) A fixed extended assignment maps all variables to six domain groups organised around the SOFA organ systems.%
    }
    \label{fig:architecture}
\end{figure*}

SepsisLens has three stages: observation-aware variable representations,
variable-preserving temporal encoding, and structured risk composition.
The design keeps variables identifiable until risk composition, so the
clinical component path remains inside the forward computation and is not added after prediction. Figure~\ref{fig:architecture} gives an
architecture overview.

\subsection{Observation-Aware Variable Representations}

The first stage encodes irregular measurements as variable-specific
representations while retaining observation information
(Figure~\ref{fig:architecture}(a)). For variable $v$ at timestep $t$,
SepsisLens constructs
\begin{equation}
\mathbf{r}_{t,v}
=
[x_{t,v}, \nabla x_{t,v}, \Delta t_{t,v}, n_{t,v}],
\end{equation}
where $x_{t,v}$ is the causally carried-forward normalised value,
$\nabla x_{t,v}$ is the local trend relative to the most recent
in-ICU observation, $\Delta t_{t,v}$ is elapsed time since the
most recent in-ICU observation with zero at an observed timestep,
and $n_{t,v}$ indicates the absence of both a current observation
and a prior in-ICU observation. Pre-ICU measurements can still
enter through the carried-forward value and cumulative
observation-history mask.

A variable encoder combines the observation-aware channels with learned
variable identity and organ-domain embeddings:
\begin{equation}
\mathbf{z}_{t,v}
=
\mathrm{Proj}
\left(
[\mathbf{r}_{t,v};\mathbf{e}^{\mathrm{var}}_v;\mathbf{e}^{\mathrm{org}}_{o(v)}]
\right),
\end{equation}
where $o(v)$ denotes a fixed extended assignment that maps both
SOFA-defining variables and additional clinically related variables to six
domain groups organised around the SOFA organ systems, giving
$\mathbf{Z}\in\mathbb{R}^{B\times T\times V\times d}$.

\subsection{Variable-Preserving Temporal Encoding}

Many clinical time-series models first mix variables into a patient-level
state at each timestep. SepsisLens instead keeps the variable axis explicit.
The embedded tensor $\mathbf{Z}$ is reshaped into $BV$ separate variable
trajectories, processed by a shared temporal encoder, and reshaped back:
\begin{equation}
\mathbf{H}
=
\mathrm{reshape}
\!\left(
F_{\theta}
\!\left(
\mathrm{reshape}(\mathbf{Z})
\right)\right),
\end{equation}
where $\mathbf{H}\in\mathbb{R}^{B\times T\times V\times d}$. We implement
$F_{\theta}$ as two Mamba2 layers with RMSNorm and residual
connections~\cite{mamba,mamba2}. The same encoder parameters are applied to
every variable trajectory, which keeps the model compact.
The hidden states are not pooled over the variable dimension, so
$\mathbf{H}_{t,v}$ remains a temporally contextualised state for variable
$v$. Information from different variables is combined only during risk
composition by organ aggregation and a global residual correction.

\subsection{Structured Risk Composition}

For each horizon $h\in\mathcal{H}$, StructuredRiskHead maps the preserved
states to a risk logit using a structured component path and a global residual
correction path (Figure~\ref{fig:architecture}(c)). Let
$\widetilde{\mathbf{H}}_{t,v}=\mathrm{LayerNorm}(\mathbf{H}_{t,v})$. The
component path produces a signed variable-level score, a gate, and a gated
component:
\begin{align}
a_{t,v}^{(h)} &= \mathrm{ScoreMLP}_{h}(\widetilde{\mathbf{H}}_{t,v}),\\
g_{t,v}^{(h)} &= \sigma\left(\mathrm{GateMLP}_{h}([\widetilde{\mathbf{H}}_{t,v};M_{t,v};\Delta t_{t,v}])\right),\\
c_{t,v}^{(h)} &= g_{t,v}^{(h)}a_{t,v}^{(h)}.
\end{align}
The signed score can increase or decrease the risk logit, while the gate
controls contribution strength using the state, cumulative observation-history
mask, and elapsed time.

Variable-level components are aggregated within six fixed
domain groups organised around the SOFA systems
\cite{sofa}:
\begin{equation}
c_{t,o}^{(h)}
=
\sum_{v:o(v)=o} c_{t,v}^{(h)},
\qquad
o=1,\ldots,O,
\end{equation}
where $O=6$. This produces one component for each organ domain. These
components are computed inside the forward pass, not by post-hoc attribution.

The residual path forms a global correction from the mean-pooled variable
states:
\begin{equation}
\rho_t^{(h)}
=
\mathrm{ResidualMLP}_{h}
\left(
\frac{1}{V}\sum_{v=1}^{V}\widetilde{\mathbf{H}}_{t,v}
\right).
\end{equation}
The final risk logit is
\begin{equation}
\ell^{(h)}_t
=
\sum_{o=1}^{O}c_{t,o}^{(h)}
+
\alpha\,\rho_t^{(h)},
\qquad
\hat{y}^{(h)}_t
=
\sigma(\ell^{(h)}_t),
\end{equation}
where $\alpha$ is a scalar scale factor. The sum of the organ-level
components forms the structured contribution to the logit, while
$\alpha\rho_t^{(h)}$ provides a global residual correction.

\subsection{Training Objective}

The main objective trains multi-horizon sepsis risk over pre-onset timesteps:
\begin{equation}
\mathcal{L}_{pred}
=
\sum_{t\in\Omega}
\sum_{h\in\mathcal{H}}
w_h\,\omega_{t,h}^{+}\,\mathrm{FL}_{\gamma}
\!\left(
y_t^{(h)},\hat{y}_t^{(h)}
\right),
\end{equation}
where $\Omega$ is the set of pre-onset timesteps, $w_h$ is a horizon weight,
$\omega_{t,h}^{+}$ is the horizon-specific positive-timestep lead-time weight, and
$\mathrm{FL}_{\gamma}$ is focal binary cross-entropy with
$\gamma=2$ and a positive-class weight~\cite{lin2017focal}. Positive
pre-onset timesteps are additionally weighted to encourage earlier detection.

During training, time-aligned SOFA-derived dysfunction labels supervise the organ-level components. For ICU timesteps, let $q_{t,o}$ denote the dysfunction label for
organ domain $o$, computed from measurements available up to
$t$. Synthetic pre-ICU bins use zero-coded auxiliary targets. Using the same pre-onset timestep set as the sepsis-risk loss, the primary-horizon organ loss is
\begin{equation}
\mathcal{L}_{organ}
=
\sum_{t\in\Omega}
\sum_{o=1}^{O}
\mathrm{BCE}
\!\left(
q_{t,o},\,
\sigma(c_{t,o}^{(6)})
\right).
\end{equation}
These labels supervise only the organ-level components during training and are never used as model inputs. At inference, SepsisLens receives only the patient trajectory. For an organ domain without label construction, we set the corresponding
$q_{t,o}$ to zero, so the domain receives only negative auxiliary targets.
Such domains are not interpreted as supervised organ-dysfunction estimates.

An $L_1$ sparsity term
$\mathcal{L}_{sparse}=\sum_{t\in\Omega}\sum_{v=1}^{V}|a_{t,v}^{(6)}|$
on primary-horizon variable scores discourages diffuse variable-level scores when $\lambda_{sp}>0$. The full objective is
\begin{equation}
\mathcal{L}
=
\mathcal{L}_{pred}
+
\lambda_{org}\mathcal{L}_{organ}
+
\lambda_{sp}\mathcal{L}_{sparse}.
\end{equation}
The displayed sums omit normalisation constants. At inference, SepsisLens
receives only the patient trajectory and returns multi-horizon risk scores
together with variable-level and organ-level components.

\section{Experimental Setup}

\subsection{Datasets}

We evaluate SepsisLens on four cohorts that cover different early-warning settings. All cohorts are converted to the same hourly format and use the same 31 clinical variables when available. MIMIC-IV \cite{johnson2023mimic} is the primary cohort, with 66,237 ICU admissions split into 53,057 training, 6,602 validation, and 6,578 test admissions. MIMIC-III \cite{johnson2016mimic} provides an earlier public ICU cohort, with 46,676 ICU admissions split into 37,327 training, 4,689 validation, and 4,660 test admissions. eICU \cite{pollard2018eicu} is a multi-centre benchmark with 22,670 ICU unit stays split into 18,141 training, 2,278 validation, and 2,251 test unit stays, with 20\% sepsis prevalence under a 1:4 case-control benchmark design. The private-hospital cohort contains 38,813 ICU admissions with 9.4\% sepsis prevalence, split into 27,169 training, 5,821 validation, and 5,823 test admissions. This independent hospital cohort uses site-specific training, validation, and test splits. It has reduced feature availability, with 11 of the 31 variables unavailable after feature harmonisation.

\subsection{Baselines}
We compare SepsisLens with learning-based baselines and fixed-rule clinical scores. The learning-based models are LSTM, GRU, GRU-D \cite{grud}, Transformer, Raindrop \cite{raindrop}, RETAIN \cite{choi2016retain}, and SepsisCalc \cite{sepsiscalc}. The clinical scores are NEWS, MEWS, qSOFA, and SIRS. For the primary comparison, all learning-based models use the same
patient-level splits, input variables, primary 6\,h evaluation target,
and validation-selected threshold protocol. The baseline models are
trained with binary cross-entropy and validation-based early stopping. Clinical scores use the available variables required by each score. All four scores have AUROC\,$<.61$ in all cohorts, well below the learning-based models.

\subsection{Metrics and Threshold Protocol}
We report AUROC and AUPRC as threshold-independent discrimination
metrics, denoted AUC and APC in the tables. We also report F1 and
operational metrics: event recall (eR: the fraction of sepsis stays
with at least one pre-onset alert), false alarm rate (FAR: the mean
fraction of alerting timesteps among non-sepsis stays), and median
lead time (ldT: the time from first pre-onset alert to onset among
detected events). Alerts may occur at any pre-onset timestep, allowing
ldT to exceed the 6\,h label window. For each model and seed, the primary 6\,h threshold is selected on the
validation set by maximising timestep-level F1, then applied once to
the corresponding held-out test predictions. Matched-recall operating
points are derived from threshold sweeps; structural ablations use a
fixed threshold of 0.5. The masking flip metric uses the
validation-selected alert threshold.

All learning-based experiments use three seeds (42, 3407, 2026).
Results are mean$\,\pm\,$std unless stated otherwise. Normalisation
statistics are estimated from the training split only. Timesteps at or
after sepsis onset are excluded from training loss, threshold selection,
and test evaluation.

\section{Results}

The evaluation first reports predictive and operational performance on
four cohorts, then examines alert burden, the contribution of the
structural design, and the functional relevance of the risk components.

\subsection{Q1: Primary 6-Hour Early-Warning Results}

Table~\ref{tab:mimic4} compares SepsisLens with seven baselines under
the common pre-onset protocol. SepsisLens achieves the highest AUC,
APC, and F1 and the lowest FAR with 91.1K parameters. Although
SepsisCalc, GRU-D, and Raindrop reach higher event recall at their
validation-selected thresholds, their FARs are 1.8--2.1 times higher.
SepsisLens also requires the lowest FAR at matched event recall
(Table~\ref{tab:fixed_evtr}), while median lead time remains comparable
among models.
\begin{table}[t]
\centering
{\small

{
\setlength{\tabcolsep}{3pt}
\begin{tabular*}{\columnwidth}
{@{\extracolsep{\fill}}lrrrr@{}}
\toprule
Method & \#P (K) & AUC & APC & F1 \\
\midrule
SepsisLens
  & \textbf{91.1}
  & $\mathbf{.939{\pm}.004}$
  & $\mathbf{.474{\pm}.019}$
  & $\mathbf{.504{\pm}.017}$ \\
SepsisCalc
  & 225.0
  & $.910{\pm}.001$
  & $.271{\pm}.005$
  & $.396{\pm}.002$ \\
Transformer
  & 400.8
  & $.909{\pm}.001$
  & $.293{\pm}.006$
  & $.407{\pm}.003$ \\
GRU-D
  & 222.1
  & $.905{\pm}.002$
  & $.299{\pm}.009$
  & $.418{\pm}.006$ \\
LSTM
  & 214.7
  & $.904{\pm}.000$
  & $.288{\pm}.004$
  & $.409{\pm}.002$ \\
Raindrop
  & 235.4
  & $.904{\pm}.002$
  & $.268{\pm}.009$
  & $.396{\pm}.006$ \\
GRU
  & 161.0
  & $.899{\pm}.002$
  & $.270{\pm}.010$
  & $.397{\pm}.005$ \\
RETAIN
  & 206.4
  & $.802{\pm}.001$
  & $.214{\pm}.006$
  & $.305{\pm}.007$ \\
\bottomrule
\end{tabular*}
}

\vspace{3pt}

\begin{tabular*}{\columnwidth}
{@{\extracolsep{\fill}}lrrr@{}}
\toprule
Method & eR & FAR & ldT \\
\midrule
SepsisLens
  & $.831{\pm}.014$
  & $\mathbf{.046{\pm}.007}$
  & $5.6{\pm}0.1$ \\
SepsisCalc
  & $.941{\pm}.004$
  & $.097{\pm}.005$
  & $5.5{\pm}0.1$ \\
Transformer
  & $.882{\pm}.011$
  & $.084{\pm}.004$
  & $5.7{\pm}0.1$ \\
GRU-D
  & $.924{\pm}.007$
  & $.082{\pm}.003$
  & $5.5{\pm}0.1$ \\
LSTM
  & $.904{\pm}.017$
  & $.085{\pm}.004$
  & $5.7{\pm}0.1$ \\
Raindrop
  & $.924{\pm}.007$
  & $.096{\pm}.004$
  & $5.7{\pm}0.1$ \\
GRU
  & $.910{\pm}.006$
  & $.090{\pm}.003$
  & $5.6{\pm}0.0$ \\
RETAIN
  & $.705{\pm}.032$
  & $.091{\pm}.009$
  & $4.9{\pm}0.3$ \\
\bottomrule
\end{tabular*}
}

\caption{Primary 6\,h early-warning results on MIMIC-IV
(mean$\,\pm\,$std over three seeds).}
\label{tab:mimic4}
\end{table}

\subsection{Q2: Does the Design Remain Effective Across Cohorts?}

Table~\ref{tab:cohorts} evaluates SepsisLens on four ICU cohorts.
It maintains strong AUC and APC on all four cohorts. The eICU
operating point is conservative, with low FAR and lower event recall.
On the private cohort, where eleven variables are unavailable after
harmonisation, SepsisLens reaches an AUC of $.973$ and FAR of $.019$.
MIMIC-III has FAR below $.001$ before rounding and a median lead time
of $6.3$\,h. All models show unusually high discrimination, suggesting
a cohort effect under this event definition.

\begin{table}[t]
\centering
{\small

\begin{tabular*}{\columnwidth}
{@{\extracolsep{\fill}}lrrrr@{}}
\toprule
Cohort & Test $n$ & AUC & APC & F1 \\
\midrule
MIMIC-IV
  & 6{,}578
  & $.939{\pm}.004$
  & $.474{\pm}.019$
  & $.504{\pm}.017$ \\
MIMIC-III
  & 4{,}660
  & $.996{\pm}.000$
  & $.763{\pm}.005$
  & $.747{\pm}.002$ \\
eICU
  & 2{,}251
  & $.948{\pm}.002$
  & $.219{\pm}.007$
  & $.287{\pm}.004$ \\
Private
  & 5{,}823
  & $.973{\pm}.001$
  & $.404{\pm}.006$
  & $.428{\pm}.002$ \\
\bottomrule
\end{tabular*}

\vspace{3pt}

\begin{tabular*}{\columnwidth}
{@{\extracolsep{\fill}}lrrr@{}}
\toprule
Cohort & eR & FAR & ldT \\
\midrule
MIMIC-IV
  & $.831{\pm}.014$
  & $.046{\pm}.007$
  & $5.6{\pm}0.1$ \\
MIMIC-III
  & $.984{\pm}.004$
  & $.000{\pm}.000$
  & $6.3{\pm}0.1$ \\
eICU
  & $.580{\pm}.076$
  & $.009{\pm}.003$
  & $6.0{\pm}0.0$ \\
Private
  & $.641{\pm}.038$
  & $.019{\pm}.004$
  & $6.0{\pm}0.0$ \\
\bottomrule
\end{tabular*}
}

\caption{SepsisLens 6\,h early-warning results on ICU cohorts.}
\label{tab:cohorts}
\end{table}

\subsection{Q3: Does SepsisLens Reduce Alert Burden?}

\begin{figure*}[t]
\centering
\includegraphics[width=\textwidth]{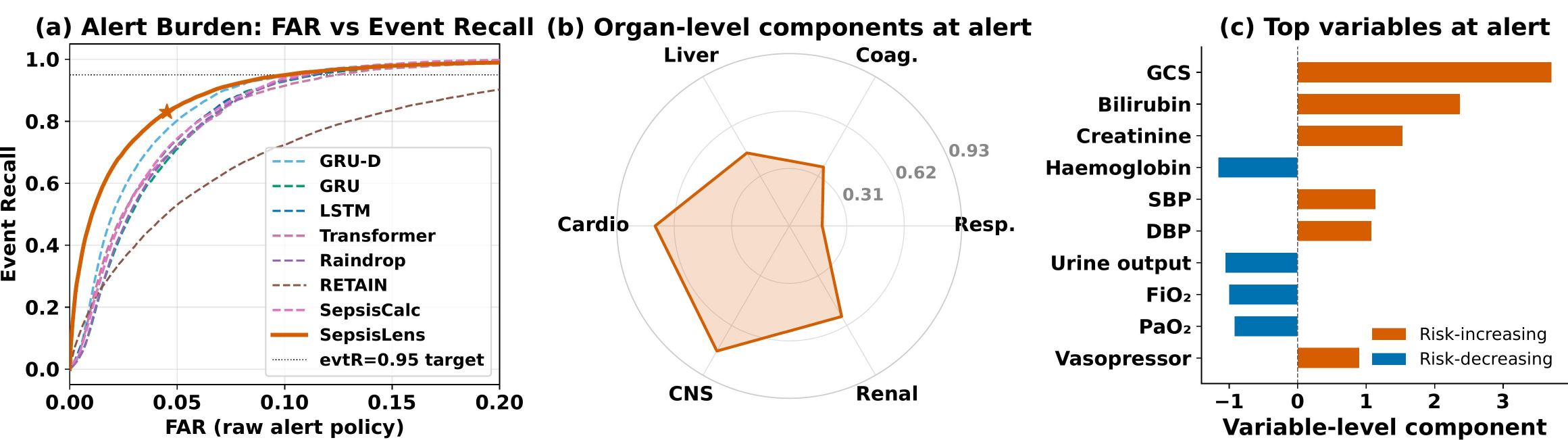}
\caption{Alert burden and structured risk outputs.
(a)~FAR versus event recall on MIMIC-IV over a threshold sweep. The marker denotes SepsisLens at its validation-selected threshold. The horizontal dashed line marks the $\mathrm{eR} = 0.95$ target used in Table~\ref{tab:fixed_evtr}.
(b,c)~Organ-level and variable-level components for a representative true-positive pre-onset alert. The StructuredRiskHead produces these components from variable-indexed temporal states without post-hoc attribution.}
\label{fig:cases}
\end{figure*}

Figure~\ref{fig:cases}(a) and Table~\ref{tab:fixed_evtr} evaluate alert
burden beyond a single validation-selected threshold. SepsisLens has
the lowest mean FAR over three seeds at all three event-recall targets
(.068, .083, and .100), followed by GRU-D (.072, .084, and .102).
The full sweep shows the FAR--event-recall trade-off.

\begin{table}[t]
\centering
{\small

\setlength{\tabcolsep}{2pt}

\begin{tabular}{@{}lrrr@{\hspace{6pt}}lrrr@{}}
\toprule
Method
& ${\geq}$.90
& ${\geq}$.93
& ${\geq}$.95
& Method
& ${\geq}$.90
& ${\geq}$.93
& ${\geq}$.95 \\
\midrule
SepsisLens
& \textbf{.068}
& \textbf{.083}
& \textbf{.100}
& Raindrop
& .087
& .098
& .112 \\

GRU-D
& .072
& .084
& .102
& GRU
& .086
& .102
& .113 \\

SepsisCalc
& .085
& .096
& .109
& Transf.
& .092
& .108
& .122 \\

LSTM
& .082
& .096
& .110
& RETAIN
& .196
& .223
& .250 \\
\bottomrule
\end{tabular}
}

\caption{Mean FAR over three seeds at fixed event-recall targets on MIMIC-IV. Lower is better.}
\label{tab:fixed_evtr}
\end{table}

\subsection{Q4: Which Design Choices Matter?}

Table~\ref{tab:ablation} isolates pre-ICU alignment, structured risk
composition, variable-preserving encoding, and organ supervision at a
fixed threshold of 0.5. Removing pre-ICU alignment causes the largest
degradation: AUC falls from .939 to .909, FAR rises from .223 to .298,
and median lead time falls from 7.2\,h to 1.8\,h. PlainHead and
EarlyFusion remove the structured outputs or explicit variable axis
while retaining similar scalar discrimination; their lower FAR at
0.5 reflects score distributions rather than superior operating
profiles. Removing organ supervision leaves scalar AUC nearly unchanged but
reduces oAUC from .940 to .567, showing that the auxiliary objective
aligns the organ-level components. This metric measures alignment
with the auxiliary SOFA-derived organ targets, not scalar sepsis
discrimination.

\begin{table}[t]
\centering
{\small

\begin{tabular*}{\columnwidth}
{@{\extracolsep{\fill}}lrrr@{}}
\toprule
Variant & AUC & APC & oAUC \\
\midrule
SepsisLens
  & $.939{\pm}.004$
  & $\mathbf{.474{\pm}.019}$
  & $\mathbf{.940{\pm}.013}$ \\
$-$OrgSupv
  & $.940{\pm}.002$
  & $.448{\pm}.005$
  & $.567{\pm}.015$ \\
ICUOnly
  & $.909{\pm}.002$
  & $.452{\pm}.001$
  & $.899{\pm}.003$ \\
PlainHead
  & $.935{\pm}.001$
  & $.450{\pm}.008$
  & --- \\
EarlyFusion
  & $.944{\pm}.001$
  & $.464{\pm}.017$
  & --- \\
\bottomrule
\end{tabular*}

\vspace{3pt}

\setlength{\tabcolsep}{3pt}
\begin{tabular*}{\columnwidth}
{@{\extracolsep{\fill}}lrrrr@{}}
\toprule
Variant & F1 & eR & FAR & ldT \\
\midrule
SepsisLens
  & $.330{\pm}.053$
  & $\mathbf{.991{\pm}.003}$
  & $.223{\pm}.070$
  & $7.2{\pm}0.2$ \\
$-$OrgSupv
  & $.381{\pm}.014$
  & $.989{\pm}.001$
  & $.160{\pm}.012$
  & $6.9{\pm}0.1$ \\
ICUOnly
  & $.230{\pm}.008$
  & $.986{\pm}.003$
  & $.298{\pm}.011$
  & $1.8{\pm}0.0$ \\
PlainHead
  & $.347{\pm}.036$
  & $.991{\pm}.004$
  & $.199{\pm}.043$
  & $7.1{\pm}0.1$ \\
EarlyFusion
  & $.373{\pm}.055$
  & $.988{\pm}.005$
  & $.178{\pm}.051$
  & $6.9{\pm}0.1$ \\
\bottomrule
\end{tabular*}

}

\caption{MIMIC-IV structural ablations at a fixed threshold of 0.5.
oAUC denotes organ macro-AUROC.}
\label{tab:ablation}
\end{table}

\subsection{Q5: Are Structured Risk Components Functionally Tied to Prediction?}

Figure~\ref{fig:cases}(b,c) shows the organ- and variable-level components for a representative true-positive pre-onset alert. We test functional relevance by ranking variables by positive
variable-level components at each patient's first true-positive alert,
setting top-$k$, random-$k$, or bottom-$k$ histories to the
never-observed state, and rerunning the model. Table~\ref{tab:topk} reports $\Delta p=p_\mathrm{orig}-p_\mathrm{masked}$ and alert flip rates for $k\in\{1,3,5\}$ over three seeds; random-$k$ uses 20 samples per alert. Top-ranked masking has the largest effect at every $k$, with the top five variables flipping $95.6$\% of alerts below threshold ($.956{\pm}.021$). Random masking has a smaller effect, while bottom-ranked masking produces the smallest probability drops and flip rates. The consistent top $>$ random $>$ bottom ordering indicates that the components rank variables by functional relevance to prediction, not only by their additive position in the output head.

\begin{table}[t]
\centering
{\small
\begin{tabular}{@{}llccc@{}}
\toprule
Masking & Metric & $k=1$ & $k=3$ & $k=5$ \\
\midrule
Top-$k$
& $\Delta p$ & $.066{\pm}.006$ & $.176{\pm}.030$ & $.252{\pm}.034$ \\

& Flip       & $.593{\pm}.028$ & $.884{\pm}.037$ & $.956{\pm}.021$ \\
Random-$k$
& $\Delta p$ & $.029{\pm}.006$ & $.095{\pm}.014$ & $.195{\pm}.018$ \\

& Flip       & $.364{\pm}.030$ & $.630{\pm}.045$ & $.843{\pm}.046$ \\
Bottom-$k$
& $\Delta p$ & $.007{\pm}.014$ & $.021{\pm}.023$ & $.037{\pm}.028$ \\

& Flip       & $.180{\pm}.115$ & $.304{\pm}.154$ & $.407{\pm}.180$ \\
\bottomrule
\end{tabular}
}
\caption{Input-side variable masking on MIMIC-IV.
Flip uses the validation-selected threshold.}
\label{tab:topk}
\end{table}

\section{Discussion}

\textbf{Operational metrics change the model comparison.}
AUROC alone is insufficient for early sepsis warning. On MIMIC-IV, ranking metrics do not determine operating profiles after thresholding. Higher event recall can require many more false alarms, and lower FAR can miss more events. Reporting event recall, lead time, and FAR alongside discrimination connects model selection to alert burden and timely intervention.

\noindent \textbf{Prediction structure determines how risk is composed.}
SepsisLens preserves clinical structure before the risk score is computed. Post-hoc attribution can inspect a completed score but does not determine its composition. The ablations separate the roles of the structured head and variable-preserving encoding. PlainHead removes the structured component outputs and slightly reduces AUROC. EarlyFusion removes the explicit variable axis before temporal modelling while keeping scalar AUROC similar. These results support the central design choice: preserve variable-level states until risk composition, then form a structured contribution from variable-level and organ-level components. This design keeps the component path available without relying on a larger model. The compact parameter count shows that the design does not require a large model.

\noindent \textbf{Organ supervision aligns the organ-level components.}
The $-$OrgSupv ablation separates scalar prediction from component supervision. Removing organ supervision leaves scalar AUROC close to the full model (.939$\to$.940), but reduces organ macro-AUROC from .940 to .567. This result should be read as supervised alignment of organ-level components with time-aligned SOFA-derived labels, not as an unsupervised discovery of organ dysfunction. The labels are derived only from measurements available up to each prediction timestep and are not used as model inputs. SepsisLens uses this auxiliary objective to align the organ-level components while keeping inference trajectory-only.

\noindent \textbf{Limitations.}
This study is retrospective. Prospective deployment would require site-specific alert policies, calibration monitoring, and clinician-facing evaluation. The private-hospital cohort has incomplete organ labels, and MIMIC-III appears unusually easy under this event definition, so neither should be treated as standalone evidence for deployment readiness. The structured components are part of the forward computation, but they are supervised risk components, not causal explanations or independently discovered organ states.

\section{Conclusion}

SepsisLens formulates early sepsis warning as a structure-preserving sequence modelling problem. It preserves variable-indexed temporal states and forms multi-horizon risk using an explicit variable-level and organ-level component path with a global residual correction. On four ICU cohorts, SepsisLens shows strong discrimination, while MIMIC-IV results show lower alert burden at matched event recall. Structural ablations and input-side masking further support the proposed design. These results support early-warning models that preserve clinical structure during prediction instead of adding it after prediction.

\bibliography{sepsislens_refs}

@article{sepsis3,
  title        = {The Third International Consensus Definitions for Sepsis and Septic Shock ({Sepsis-3})},
  author       = {Singer, Mervyn and Deutschman, Clifford S. and Seymour, Christopher W. and Shankar-Hari, Manu and Annane, Djillali and Bauer, Michael and Bellomo, Rinaldo and Bernard, Gordon R. and Chiche, Jean-Daniel and Coopersmith, Craig M. and Hotchkiss, Richard S. and Levy, Mitchell M. and Marshall, John C. and Martin, Greg S. and Opal, Steven M. and Rubenfeld, Gordon D. and van der Poll, Tom and Vincent, Jean-Louis and Angus, Derek C.},
  journal      = {JAMA},
  volume       = {315},
  number       = {8},
  pages        = {801--810},
  year         = {2016},
  doi          = {10.1001/jama.2016.0287}
}

@article{sofa,
  title        = {The {SOFA} ({Sepsis}-related Organ Failure Assessment) Score to Describe Organ Dysfunction/Failure},
  author       = {Vincent, Jean-Louis and Moreno, Rui and Takala, Jukka and Willatts, Sheila and de Mendon{\c{c}}a, Antonio and Bruining, Hans and Reinhart, Konrad and Suter, Peter M. and Thijs, L. G.},
  journal      = {Intensive Care Medicine},
  volume       = {22},
  number       = {7},
  pages        = {707--710},
  year         = {1996},
  doi          = {10.1007/BF01709751}
}

@article{grud,
  title        = {Recurrent Neural Networks for Multivariate Time Series with Missing Values},
  author       = {Che, Zhengping and Purushotham, Sanjay and Cho, Kyunghyun and Sontag, David and Liu, Yan},
  journal      = {Scientific Reports},
  volume       = {8},
  number       = {1},
  pages        = {6085},
  year         = {2018},
  doi          = {10.1038/s41598-018-24271-9}
}

@inproceedings{choi2016retain,
  title        = {{RETAIN}: An Interpretable Predictive Model for Healthcare Using Reverse Time Attention Mechanism},
  author       = {Choi, Edward and Bahadori, Mohammad Taha and Sun, Jimeng and Kulas, Joshua and Schuetz, Andy and Stewart, Walter},
  booktitle    = {Advances in Neural Information Processing Systems},
  volume       = {29},
  pages        = {3504--3512},
  year         = {2016}
}

@inproceedings{futoma2017mgprnn,
  title        = {Learning to Detect Sepsis with a Multitask {Gaussian} Process {RNN} Classifier},
  author       = {Futoma, Joseph and Hariharan, Sanjay and Heller, Katherine},
  booktitle    = {Proceedings of the 34th International Conference on Machine Learning},
  series       = {Proceedings of Machine Learning Research},
  volume       = {70},
  pages        = {1174--1182},
  publisher    = {PMLR},
  year         = {2017}
}

@article{rosnati2021mgpatttcn,
  title        = {{MGP-AttTCN}: An Interpretable Machine Learning Model for the Prediction of Sepsis},
  author       = {Rosnati, Margherita and Fortuin, Vincent},
  journal      = {PLOS ONE},
  volume       = {16},
  number       = {5},
  pages        = {e0251248},
  year         = {2021},
  doi          = {10.1371/journal.pone.0251248}
}

@inproceedings{shukla2021mtan,
  title        = {Multi-Time Attention Networks for Irregularly Sampled Time Series},
  author       = {Shukla, Satya Narayan and Marlin, Benjamin M.},
  booktitle    = {International Conference on Learning Representations},
  year         = {2021}
}

@inproceedings{raindrop,
  title        = {Graph-Guided Network for Irregularly Sampled Multivariate Time Series},
  author       = {Zhang, Xiang and Zeman, Marko and Tsiligkaridis, Theodoros and Zitnik, Marinka},
  booktitle    = {International Conference on Learning Representations},
  year         = {2022}
}

@inproceedings{sepsiscalc,
  title        = {{SepsisCalc}: Integrating Clinical Calculators into Early Sepsis Prediction via Dynamic Temporal Graph},
  author       = {Yin, Changchang and Fu, Shihan and Yao, Bingsheng and Pham, Thai-Hoang and Cao, Weidan and Wang, Dakuo and Caterino, Jeffrey and Zhang, Ping},
  booktitle    = {Proceedings of the 31st {ACM} {SIGKDD} Conference on Knowledge Discovery and Data Mining},
  pages        = {2779--2790},
  year         = {2025},
  publisher    = {Association for Computing Machinery},
  doi          = {10.1145/3690624.3709402}
}

@inproceedings{jain2019attention,
  title        = {Attention is not Explanation},
  author       = {Jain, Sarthak and Wallace, Byron C.},
  booktitle    = {Proceedings of the 2019 Conference of the North American Chapter of the Association for Computational Linguistics: Human Language Technologies},
  pages        = {3543--3556},
  year         = {2019},
  publisher    = {Association for Computational Linguistics},
  doi          = {10.18653/v1/N19-1357}
}

@article{rudin2019stop,
  title        = {Stop Explaining Black Box Machine Learning Models for High Stakes Decisions and Use Interpretable Models Instead},
  author       = {Rudin, Cynthia},
  journal      = {Nature Machine Intelligence},
  volume       = {1},
  pages        = {206--215},
  year         = {2019},
  doi          = {10.1038/s42256-019-0048-x}
}

@misc{mamba,
  title        = {Mamba: Linear-Time Sequence Modeling with Selective State Spaces},
  author       = {Gu, Albert and Dao, Tri},
  year         = {2023},
  eprint       = {2312.00752},
  archivePrefix = {arXiv},
  primaryClass = {cs.LG}
}

@misc{mamba2,
  title        = {Transformers are {SSM}s: Generalized Models and Efficient Algorithms Through Structured State Space Duality},
  author       = {Dao, Tri and Gu, Albert},
  year         = {2024},
  eprint       = {2405.21060},
  archivePrefix = {arXiv},
  primaryClass = {cs.LG}
}

@article{johnson2016mimic,
  title        = {{MIMIC-III}, a Freely Accessible Critical Care Database},
  author       = {Johnson, Alistair E. W. and Pollard, Tom J. and Shen, Lu and Lehman, Li-Wei H. and Feng, Mengling and Ghassemi, Mohammad and Moody, Benjamin and Szolovits, Peter and Celi, Leo Anthony and Mark, Roger G.},
  journal      = {Scientific Data},
  volume       = {3},
  pages        = {160035},
  year         = {2016},
  doi          = {10.1038/sdata.2016.35}
}

@article{johnson2023mimic,
  title        = {{MIMIC-IV}, a Freely Accessible Electronic Health Record Dataset},
  author       = {Johnson, Alistair E. W. and Bulgarelli, Lucas and Shen, Lu and Gayles, Alvin and Shammout, Ayad and Horng, Steven and Pollard, Tom J. and Hao, Sicheng and Moody, Benjamin and Gow, Brian and Lehman, Li-Wei H. and Celi, Leo Anthony and Mark, Roger G.},
  journal      = {Scientific Data},
  volume       = {10},
  number       = {1},
  pages        = {1},
  year         = {2023},
  doi          = {10.1038/s41597-022-01899-x}
}

@article{pollard2018eicu,
  title        = {The {eICU} Collaborative Research Database, a Freely Available Multi-Center Database for Critical Care Research},
  author       = {Pollard, Tom J. and Johnson, Alistair E. W. and Raffa, Jesse D. and Celi, Leo Anthony and Mark, Roger G. and Badawi, Omar},
  journal      = {Scientific Data},
  volume       = {5},
  pages        = {180178},
  year         = {2018},
  doi          = {10.1038/sdata.2018.178}
}

@inproceedings{lin2017focal,
  title        = {Focal Loss for Dense Object Detection},
  author       = {Lin, Tsung-Yi and Goyal, Priya and Girshick, Ross and He, Kaiming and Doll{\'a}r, Piotr},
  booktitle    = {Proceedings of the IEEE International Conference on Computer Vision},
  pages        = {2980--2988},
  year         = {2017}
}

@inproceedings{xiao2025xtsformer,
  title = {{XTSFormer}: Cross-Temporal-Scale Transformer for Irregular-Time Event Prediction in Clinical Applications},
  author = {Xiao, Tingsong and Xu, Zelin and He, Wenchong and Xiao, Zhengkun and Zhang, Yupu and Liu, Zibo and Chen, Shigang and Thai, My T. and Bian, Jiang and Rashidi, Parisa and Jiang, Zhe},
  booktitle = {Proceedings of the AAAI Conference on Artificial Intelligence},
  volume = {39},
  pages = {28502--28510},
  year = {2025},
  doi = {10.1609/aaai.v39i27.35073}
}

@inproceedings{liu2025timecheat,
  title = {{TimeCHEAT}: A Channel Harmony Strategy for Irregularly Sampled Multivariate Time Series Analysis},
  author = {Liu, Jiexi and Cao, Meng and Chen, Songcan},
  booktitle = {Proceedings of the AAAI Conference on Artificial Intelligence},
  volume = {39},
  pages = {18861--18869},
  year = {2025},
  doi = {10.1609/aaai.v39i18.34076}
}

@inproceedings{liu2026apn,
  title = {Rethinking Irregular Time Series Forecasting: A Simple Yet Effective Baseline},
  author = {Liu, Xvyuan and Qiu, Xiangfei and Wu, Xingjian and Li, Zhengyu and Guo, Chenjuan and Hu, Jilin and Yang, Bin},
  booktitle = {Proceedings of the AAAI Conference on Artificial Intelligence},
  volume = {40},
  pages = {23873--23881},
  year = {2026},
  doi = {10.1609/aaai.v40i28.39563}
}

@inproceedings{zhang2024tpatchgnn,
  title = {Irregular Multivariate Time Series Forecasting: A Transformable Patching Graph Neural Networks Approach},
  author = {Zhang, Weijia and Yin, Chenlong and Liu, Hao and Zhou, Xiaofang and Xiong, Hui},
  booktitle = {Proceedings of the 41st International Conference on Machine Learning},
  series = {Proceedings of Machine Learning Research},
  volume = {235},
  pages = {60179--60196},
  year = {2024},
  publisher = {PMLR}
}

@article{wong2021external,
  title={External Validation of a Widely Implemented Proprietary Sepsis Prediction Model in Hospitalized Patients},
  author={Wong, Andrew and Otles, Erkin and Donnelly, John P. and others},
  journal={JAMA Internal Medicine},
  volume={181},
  number={8},
  pages={1065--1070},
  year={2021},
  doi={10.1001/jamainternmed.2021.2626}
}
\end{document}